\documentclass{ecai}
\usepackage{times}
\usepackage{graphicx}
\usepackage{latexsym}
\usepackage{subcaption}
\usepackage{amsmath}
\usepackage{algorithm}
\usepackage{algorithmic}

\begin{document}

\title{Towards Lossless Binary Convolutional Neural Networks Using Piecewise Approximation}

\author{Baozhou Zhu 
\institute{Delft University of Technology,The Netherlands,email:b.zhu-1@tudelft.nl} \and Zaid Al-Ars
\institute{Delft University of Technology,The Netherlands,email:z.al-ars@tudelft.nl} \and Wei Pan
\institute{Delft University of Technology,The Netherlands,email:wei.pan@tudelft.nl}
}

\maketitle
\bibliographystyle{ecai}

\begin{abstract}
  Binary Convolutional Neural Networks (CNNs) can significantly reduce the number of arithmetic operations and the size of memory storage, which makes the deployment of CNNs on mobile or embedded systems more promising. However, the accuracy degradation of single and multiple binary CNNs is unacceptable for modern architectures and large scale datasets like ImageNet. In this paper, we proposed a Piecewise Approximation (PA) scheme for multiple binary CNNs which lessens accuracy loss by approximating full precision weights and activations efficiently, and maintains parallelism of bitwise operations to guarantee efficiency. Unlike previous approaches, the proposed PA scheme segments piece-wisely the full precision weights and activations, and approximates each piece with a scaling coefficient. Our implementation on ResNet with different depths on ImageNet can reduce both Top-1 and Top-5 classification accuracy gap compared with full precision to approximately $1.0\%$. Benefited from the binarization of the downsampling layer, our proposed PA-ResNet50 requires less memory usage and two times Flops than single binary CNNs with $4$ weights and $5$ activations bases. The PA scheme can also generalize to other architectures like DenseNet and MobileNet with similar approximation power as ResNet which is promising for other tasks using binary convolutions. The code and pretrained models will be publicly available.
\end{abstract}

\section{Introduction}
CNNs have emerged as one of the most influential neural network architectures to tackle large scale machine learning problems in image recognition, natural language processing, and audio analysis \cite{he2016deep,huang2017densely}. At the same time, their deployment on mobile devices and embedded systems are gaining more and more attention due to the increasing interest from industry and academia \cite{howard2017mobilenets,sandler2018mobilenetv2}. However, the limited storage and computation resources provided by these platforms are an obstacle that is being addressed by numerous researchers working to reduce the complexity of CNNs \cite{he2018soft,zhang2015efficient,lavin2016fast,sze2017efficient}.
Fixed-point CNNs \cite{mishra2017apprentice,baozhou2019diminished,zhou2017incremental,Zhang_2018_ECCV,zhou2016dorefa,faraone2018syq} achieve even no accuracy loss with a suitable selection of bit-width, but the multiplication and the overflow processing of addition require considerable overhead. Binary CNNs have been demonstrated as a promising technique to make the deployment of CNNs feasible \cite{courbariaux2016binarized,tseng2018deterministic,darabi2018bnn+, liu2019circulant}. In single binary CNNs, full precision weights and activations are binarized into $1$ bit, so the multiplication and addition of the convolution are transformed into simple bitwise operations, resulting in significant storage and computation requirements reduction \cite{rastegari2016xnor}. The accuracy degradation of the recently enhanced single binary CNN \cite{liu2018bi} is still high ($12.9\%$ Top-1 and $9.7\%$ Top-5 accuracy degradation for ResNet18 on ImageNet) since much information has been discarded during binarization. ABC-Net \cite{lin2017towards} is the first multiple binary CNN, which shows encouraging result (around $5\%$ Top-1 and Top-5 accuracy degradation for ResNet on ImageNet). \cite{fromm2018heterogeneous,guo2017network,tang2017train,li2017performance} calculate a series of binary values and their corresponding scaling coefficients through minimizing the residual error recursively, but they can not be paralleled. \cite{zhuang2019structured} propose Group-Net to explore structure approximation, and it is a complimentary approximation to value approximation. Multiple binary CNNs can be considered as a moderate way of quantization, that is much more accurate than single binary CNNs and more efficient than fix-point CNNs. But, there is still a considerable gap between full precision implementations and multiple binary CNNs, despite the fact that an unlimited number of weights and activation bases can be used. 


To further reduce the gap between the full precision and multiple binary CNNs, we proposed Piece-wise Approximation (PA) scheme in this paper. Our main contributions are summarized as follows.
\begin{itemize}
\item PA scheme segments the whole range of the full precision weights and activations into many pieces and uses a scaling coefficient to approximate each of them, which can maintain parallelism of bitwise operation and lessen accuracy loss.
\item With less overhead, our scheme achieves much higher accuracy than ABC-Net, which indicates that it provides a better approximation for multiple binary CNNs. Benefited from the binarization of the downsampling layer, our proposed PAResNet50 requires less memory usage and two times Flops than Bi-Real Net with $4$ weights and $5$ activations
bases, which shows its potential efficiency advantage over single binary CNNs with a deeper network.
\item With the increase of the number of the weight and activation bases, our proposed PA scheme achieves the highest classification accuracy for ResNet on ImageNet among all state-of-the-art single and multiple binary CNNs.
\end{itemize}

\section{Related work}
In this Section, we describe the forward propagation and backpropagation of typical schemes to quantize CNNs. In addition, the advantages and disadvantages of these quantized CNNs are discussed concerning efficiency and accuracy.

\subsection{Single Binary Convolutional Neural Networks}
In single binary convolutional neural networks \cite{courbariaux2016binarized,tseng2018deterministic,darabi2018bnn+,shen2019searching,liu2019circulant}, weights and activations are constrained to a single value $+1$ or $-1$. The deterministic binarization function is described as follows. 
\begin{equation}
{x^b} = \left\{ \begin{array}{l}
 + 1,{x^r} \ge 0\\
 - 1,{x^r} < 0
\end{array} \right.
\end{equation}
where ${x^b}$ is the binarized variable, and ${x^r}$ is the real-valued variable. 
During the backpropagation, the “Straight-Through Estimator” (STE) method \cite{bengio2013estimating} is adapted to calculate the derivatives of the binarization functions as follows, where $C$ is the loss function. 
\begin{equation}
\frac{{\partial C}}{{\partial {x^r}}} = \frac{{\partial C}}{{\partial {x^b}}}
\end{equation}

Single binary CNNs is the most efficient quantization scheme among all the quantization schemes described in this paper. But, its accuracy degradation is too high to be deployed in practice.

\subsection{Ternary Convolutional Neural Networks}
In ternary convolutional neural networks \cite{Li2016TernaryWN,zhu2016trained,wan2018tbn,he2019simultaneously}, ternary weights are used to reduce the accuracy loss of single binary CNNs by introducing $0$ as the third quantized value, as follows.
\begin{equation}
{x^t} = \left\{ \begin{array}{l}
x^p:{x^r} > {\Delta }\\
0:\left| {{x^r}} \right| \le {\Delta }\\
 - x^n:{x^r} <  - {\Delta }
\end{array} \right.
\end{equation}
where $x^p$ and $x^n$ are the positive and negative scaling coefficients, respectively, and ${\Delta}$ is a threshold to determine the ternarized variable ${x^t}$. During the backpropagation, the STE method is still applied.

Although the introduction of $0$ improves the accuracy of single binary CNNs, it is still unacceptable to be deployed especially while training advanced CNNs on large scale dataset.

\subsection{Fixed-point Convolutional Neural Networks}
In fixed-point convolutional neural network \cite{zhou2016dorefa,Zhang_2018_ECCV,faraone2018syq,jung2019learning}, weights, activations, and gradients are quantized using fixed-point numbers of different bit-widths. Taking the weights as an example, the quantization works as follows. 
\begin{equation}
{x^f} = 2 \cdot \texttt{quantize}_f(\frac{{\tanh ({x^r})}}{{2\max (\left| {\tanh ({x^r})} \right|)}} + \frac{1}{2}) - 1
\end{equation}
where $\texttt{quantize}_f$ function quantizes the real-valued number ${x^r}$ to the $f$-bit fixed-point number ${x^f}$. During the backpropagation, the STE method still works.

With a configuration of different bit-widths for the weights, activations, and gradients, the accuracy degradation of DoReFa-Net can be preserved and controlled. But, fixed-point multipliers result in the most substantial overhead among that of all the quantization schemes in this paper. 

\subsection{Multiple Binary Convolutional Neural Networks}
In multiple binary convolutional neural networks \cite{lin2017towards,fromm2018heterogeneous,guo2017network,tang2017train,li2017performance,zhuang2019structured}, a combination of multiple binary bases is adopted to approximate full precision weights and activations. Following is the weights approximation using linear combination.
\begin{equation}
{x^r} = \sum\limits_{i = 1}^P {{\varepsilon _i}{D_i}} 
\end{equation}
where ${\varepsilon _i}$ is a trainable scaling coefficient and ${D_i}$ is a binary ($-1$ and $+1$) weight base. During the backpropagation, STE method is still used.

The adoption of multiple binary bases in ABC-Net can lessen accuracy loss compared to single binary CNNs and maintain efficiency by using parallel bitwise operations compared to fix-point CNNs. Unfortunately, there is still a considerable gap between ABC-Net and full precision although as many as needed weight and activation binary bases can be used.

\section{Piecewise approximation scheme}
In this section, the PA scheme for multiple binary CNNs is illustrated, including the approximations of weights and activations. Also, the training algorithm and the inference architecture of PA-Net are clarified. 

\begin{figure}[h]
\begin{center}
   \includegraphics[width=0.70\columnwidth]{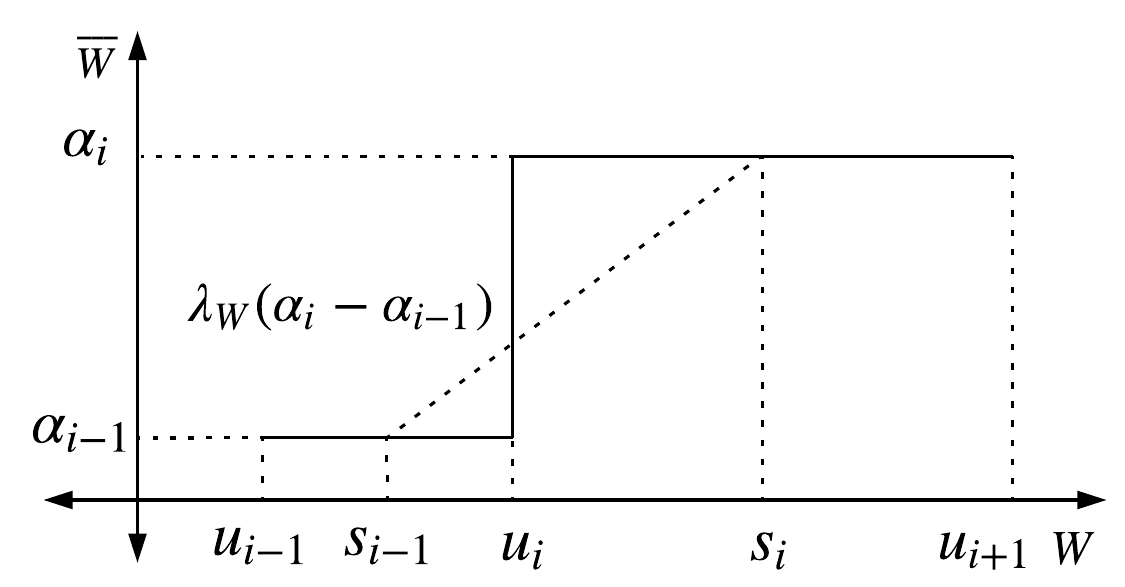}
\end{center}
   \caption{A sample of the forward propagation and backpropagation of weights approximation}
\label{figure:weights forward propagation and backpropagation}
\end{figure}
\subsection{Weights approximation}
Since approximating weights channel-wise needs much more computational resources during training, we approximated weights as a whole in this paper.

The real-valued weights are $W \in {R^{h \times w \times {c_{in}}\times{c_{out}}}}$, where $h$, $w$, ${c_{in}}$ and ${c_{out}}$ represent the height and width of a filter, the number of input and output channels, respectively. In the forward propagation of PA scheme, these are estimated by $\overline W$, which is a piecewise function composed of the following $M + 1$ pieces.
\begin{equation}
W \approx \overline W  = \left\{ \begin{array}{l}
{\alpha _1} \times {B_W},W_j \in [ - \infty ,{u_1}]\\
{\alpha _i} \times {B_W},W_j \in [{u_{i - 1}},{u_i}], i \in [2,\frac{M}{2}]\\
0.0 \times {B_W},W_j \in [{u_{\frac{M}{2}}},{u_{\frac{M}{2} + 1}}]\\
{\alpha _i} \times {B_W},W_j \in [{u_i},{u_{i + 1}}], \\
~~~~~~~~~~~~~~~~~~~~i \in [\frac{M}{2} + 1,M - 1]\\
{\alpha _M} \times {B_W},W_j \in [{u_M}, + \infty ]
\end{array} \right.
\end{equation}
where ${u_i}$ and ${\alpha _i}$ are the endpoint and scaling coefficient of the pieces, respectively. $W_j$ is a scalar and a single weight of the tensor $W$. $W_j \in [ - \infty ,{u_1}]$ refers to the $j^\text{th}$ weight of the tensor $W$ which is in the range of $[ - \infty ,{u_1}]$. ${B_W}$ is a tensor with all the values equal to $1.0$ and has the same shape as $W$. Since the distribution of the weights is close to Gaussian, all the endpoints of the weights are fixed using $mean(W)$ and $std(W)$, which refer to the mean and standard deviation of the full precision weights, respectively. The $M$ endpoints are almost uniformly sampled from $-2.0 \times std(W)$ to $2.0 \times std(W)$ except those near $0.0$. To set the endpoints of the weights properly, we attempted some different settings, where the performance difference is negligible. Taking $M = 8$ as an example, we directly recommend the endpoints set as listed in Table~\ref{table:endpoints}.

\begin{table}
\caption{Endpoints of the weights with $M = 8$}
\label{table:endpoints}
\centering
\begin{tabular}{l|l|l|l|l}
\hline
Variables & ${u_1}$ & ${u_2}$ & ${u_3}$ & ${u_4}$ \\
\hline
Values ($\times$ $std(W)$) & $-1.5$  & $-1.0$ & $-0.5$ & $-0.25$ \\
\hline
Variables & ${u_5}$ & ${u_6}$ & ${u_7}$ & ${u_8}$\\
\hline
Values ($\times$ $std(W)$)  & $0.25$  & $0.5$ & $1.0$ & $1.5$\\
\hline
\end{tabular}
\end{table}

Except for the ${(\frac{M}{2} + 1)}\emph{{-th}}$ piece, the mean of all the full precision weights of every piece serves as the optimal estimation of its scaling coefficient. 
\begin{equation}
\left\{ \begin{array}{l}
{\alpha _1} = \texttt{reduce\_mean}(W),W_j \in [ - \infty ,{u_1}]\\
{\alpha _i} = \texttt{reduce\_mean}(W),W_j \in [{u_{i - 1}},{u_i}], i \in [2,\frac{M}{2}]\\
{\alpha _i} = \texttt{reduce\_mean}(W),W_j \in [{u_i},{u_{i + 1}}],\\
~~~~~~~~~~~~~~~~~~~~~~~~~~~~~~~~~~~~~~~~~~i \in [\frac{M}{2} + 1,M - 1]\\
{\alpha _M} = \texttt{reduce\_mean}(W),W_j \in [{u_M}, + \infty ]
\end{array} \right.
\end{equation}

During the backpropagation, the relationship between $\overline W$ and $W$ has to be established, and the whole range of the weights is segmented into $M$ pieces.
\begin{equation}
\frac{{\partial \overline W }}{{\partial W}} = \left\{ \begin{array}{l}
\lambda_{W} ({\alpha _2} - {\alpha _1}),W_j \in [ - \infty ,{s_1}]\\
\lambda_{W} ({\alpha _{i + 1}} - {\alpha _i}),W_j \in [{s_{i - 1}},{s_i}], \\
~~~~~~~~~~~~~~~~~~~~~~~~~i \in [2,\frac{M}{2} - 1]\\
\lambda_{W} (0.0 - {\alpha _{\frac{M}{2}}}),W_j \in [{s_{\frac{M}{2} - 1}},{s_{\frac{M}{2}}}]\\
\lambda_{W} ({\alpha _{\frac{M}{2} + 1}} - 0.0),W_j \in [{s_{\frac{M}{2}}},{s_{\frac{M}{2} + 1}}]\\
\lambda_{W} ({\alpha _{i + 1}} - {\alpha _i}),W_j \in [{s_i},{s_{i + 1}}],\\
~~~~~~~~~~~~~~~~~~~~~~~~~i \in [\frac{M}{2} + 1,M - 2]\\
\lambda_{W} ({\alpha _M} - {\alpha _{M - 1}}),W_j \in [{s_{M - 1}}, + \infty ]
\end{array} \right.
\end{equation}
where ${s_{i}}$ is the endpoint of the pieces. $\lambda_{W}$ is a hyper-parameter, which is different when a different number of weight pieces is used. The endpoint ${s_{i}}$ can be determined simply as follows
\begin{equation}
{s_i} = ({u_{i + 1}} + {u_i})/2.0,i \in [1,M - 1]
\end{equation}

The forward propagation while $W_j \in \left[ {{u_{i - 1}},{u_{i + 1}}} \right]$ and backpropagation while $W_j \in \left[ {{s_{i - 1}},{s_i}} \right]$ are presented in Figure~\ref{figure:weights forward propagation and backpropagation}, where a linear function with slope $\lambda_{W} ({\alpha _{i}} - {\alpha _{i - 1}})$ is used to approximate the piecewise function during the backpropagation.

\subsection{Activations approximation}
\begin{figure}[h]
\begin{center}
   \includegraphics[width=0.70\columnwidth]{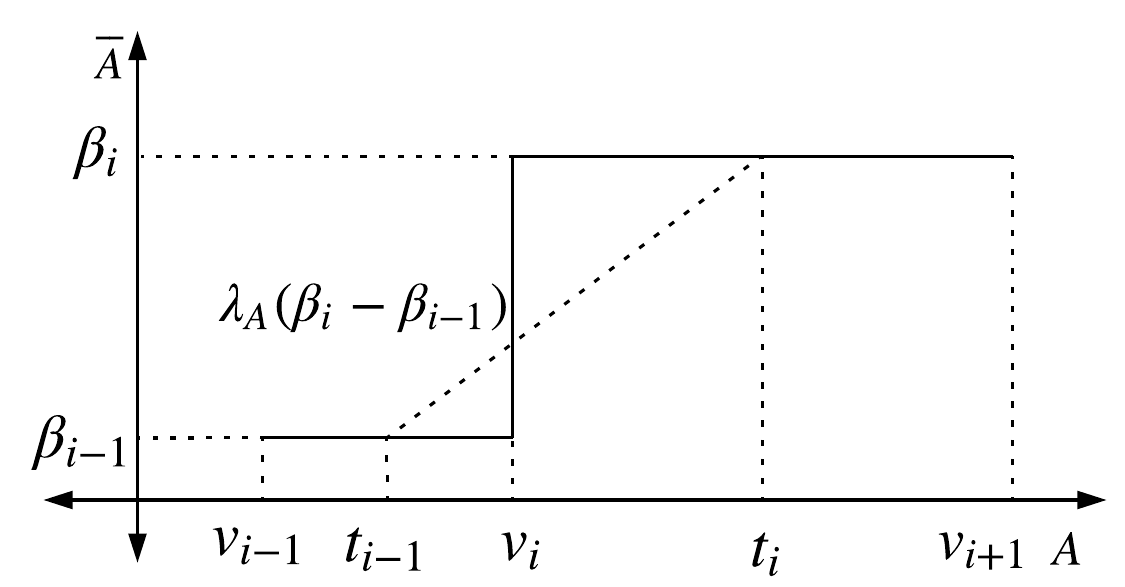}
\end{center}
   \caption{A sample of the forward propagation and backpropagation of activations approximation}
\label{figure:activations forward propagation and backpropagation}
\end{figure}
To utilize bitwise operation for convolution, activations should be binarized as well. However, the distribution of the activations will vary in the inference stage which motivates us to apply batch normalization \cite{ioffe2015batch}. Batch normalization is applied before the approximation of the activations to force them to have zero mean and unit standard deviation.

The real-valued input activations are $A \in {R^{n \times h \times w \times {c_{in}}}}$, where $n$, $h$, $w$ and ${c_{in}}$ refer to batch size, height, width and number of channels, respectively. In the forward propagation of PA scheme, these are estimated by $\overline A$, which is a piecewise function composed of the following $N + 1$ pieces.
\begin{equation}
A \approx \overline A  = \left\{ \begin{array}{l}
0.0 \times {B_A},A_j \in [ - \infty ,{v_1}]\\
{\beta _i} \times {B_A},A_j \in [{v_i},{v_{i + 1}}], i \in [1,N - 1]\\
{\beta _N} \times {B_A},A_j \in [{v_N}, + \infty ]
\end{array} \right.
\end{equation}
where ${v_{i}}$ and ${\beta _i}$ are the endpoint and scaling coefficient of the pieces, respectively. $A_j \in [{v_i},{v_{i + 1}}]$ refers to the activations of matrix $A$ which are in the closed range of $[{v_i},{v_{i + 1}}]$. ${B_A}$ is a tensor with all the values equal to $1.0$ and has the same shape as $A$. Both the endpoint ${v_i}$ and the scaling coefficient ${\beta _i}$ are trainable to learn the statistical features of the full precision activations. The bounded activation function is omitted since the endpoints are initialized with positive values. 

During the backpropagation, the relationship between $\overline A$ and $A$ has to be established, and the whole range of the activations is segmented into $N + 2$ pieces.
\begin{equation}
\frac{{\partial \overline A }}{{\partial A}} = \left\{ \begin{array}{l}
0.0,A_j \in [ - \infty ,{t_0}]\\
\lambda_{A}  \times ({\beta _1} - 0.0),A_j \in [{t_0},{t_1}]\\
\lambda_{A}  \times ({\beta _{i + 1}} - {\beta _i}),A_j \in [{t_i},{t_{i + 1}}], \\
~~~~~~~~~~~~~~~~~~~~~~~~~~~~~~~~~i = 1,...,N - 1\\
0.0,A_j \in [{t_N}, + \infty ]
\end{array} \right.
\end{equation}
where ${t_{i}}$ is the endpoint of the pieces. 
$\lambda_{A}$ is a hyper-parameter, which is the same for all the layers in a given CNN and is different between different CNNs with different depths as used this paper. The endpoint ${t_{i}}$ can be determined as follows
\begin{equation}
\left\{ \begin{array}{l}
{t_i} = ({v_i} + {v_{i + 1}})/2.0,i = 1,...,N - 1\\
{t_0} = 2.0 \times {v_0} - {s_1}\\
{t_N} = {v_N} + {\lambda _\Delta }
\end{array} \right.
\end{equation}
where ${\lambda _\Delta }$ is a hyper-parameter, which is the same for all the layers in a given CNN and is different between different CNNs in this paper.

The forward propagation while $A_j \in \left[ {{v_{i - 1}},{v_{i + 1}}} \right]$ and backpropagation while $A_j \in \left[ {{t_{i - 1}},{t_i}} \right]$ are presented in Figure~\ref{figure:activations forward propagation and backpropagation}, where a linear function with slope $\lambda_{A} ({\beta _{i}} - {\beta _{i - 1}})$ is used to approximate the piecewise function during the backpropagation.

The scaling coefficient ${\beta _i}$ is updated as follows
\begin{equation}
\begin{aligned}
&\frac{{\partial C}}{{\partial {\beta _i}}} = \frac{{\partial C}}{{\partial \overline A }}\frac{{\partial \overline A }}{{\partial {\beta _i}}} =\\
&\left\{ \begin{array}{l}
\texttt{reduce\_sum}(\frac{{\partial C}}{{\partial \overline A }}),A_j \in [{v_i},{v_{i + 1}}],i \in [1,N - 1]\\ \\
\texttt{reduce\_sum}(\frac{{\partial C}}{{\partial \overline A }}),A_j \in [{v_N}, + \infty ],i = N
\end{array} \right.
\end{aligned}
\end{equation}
Similarly, the endpoint $v_i$ is updated as
\begin{equation}
\begin{aligned}
&\frac{{\partial C}}{{\partial {v_i}}}  = \frac{{\partial C}}{{\partial \overline A }}\frac{{\partial \overline A }}{{\partial {v_i}}} =\\
&\left\{ \begin{array}{l}
{\lambda _A}({\beta _1} - 0.0) \times \texttt{reduce\_sum}(\frac{{\partial C}}{{\partial \overline A }}), \\
~~~~~~~~~~~~~~~~~~~~~~~~~~~~~~~~A_j \in [{t_0},{t_1}],i = 1\\
{\lambda _A}({\beta _i} - {\beta _{i - 1}}) \times \texttt{reduce\_sum}(\frac{{\partial C}}{{\partial \overline A }}),\\
~~~~~~~~~~~~~~~~~~~~~~~~~~~~~~~~A_j \in [{t_{i - 1}},{t_i}],i \in [2,N]
\end{array} \right.
\end{aligned}
\end{equation}

 \begin{algorithm}[ht]
 \caption{Training a $L$-layer multiple binary CNN by PA scheme}
 \label{algorithm1}
 \begin{algorithmic}[1]
 \renewcommand{\algorithmicrequire}{\textbf{Input:}}
 \renewcommand{\algorithmicensure}{\textbf{Output:}}
 \REQUIRE A mini-batch of inputs ${A_0}$ and targets ${A^ * }$, weights ${W}$. Learning rate ${\eta}$, learning rate decay factor $\lambda$. The number of endpoints $M$, scaling coefficient ${\alpha _i}$ and endpoint ${u_i}$ for weights, the number of endpoints $N$, scaling coefficient ${\beta _i}$ and endpoint ${v_i}$ for activations. PA is short for Piecewise Approximation scheme. 
 \ENSURE Updated scaling coefficient $\beta _i$, endpoint $v_i$, weights $W$ and learning rate $\eta$.
 \\  \textit{} 1. Computing the parameter gradients:
  \\ \textit{} 1.1. Forward path:
  \FOR {$k = 1$ to $L$}
  \STATE $\overline W  \leftarrow PA(W,{u_i},{\alpha _i},M)$
  \STATE $A \leftarrow Conv(\overline A ,\overline W )$
  \IF {$k < L$}
  \STATE $\overline A \leftarrow PA(A,{v_i},{\beta _i},N)$
  \ENDIF
  \ENDFOR
  \\ \textit{} 1.2. Backward propagation:
  \FOR {$k = L$ to $1$}
  \IF {$k < L$}
  \STATE $({g_A},{g_{{v_i}}},{g_{{\beta _i}}}) \leftarrow Back\_PA({g_{\overline A }},A,{v_i},{\beta _i},N)$
  \ENDIF 
  \STATE $({g_{\overline A }},{g_{\overline W }}) \leftarrow Back\_Conv({g_A},\overline A ,\overline W )$
  \STATE ${g_W} \leftarrow Back\_PA({g_{\overline W }},W,{u_i},{\alpha _i},M)$
  \ENDFOR
 \\ \textit{} 2. Accumulating the parameter gradients:
   \FOR {$k = 1$ to $L$}
  \STATE $\beta _i \leftarrow update(\beta _i,{\eta },{g_{\beta _i}})$
  \STATE $v_i \leftarrow update(v_i,{\eta},{g_{v_i}})$
  \STATE $W \leftarrow update(W,{\eta },{g_{W}})$
  \STATE ${\eta} \leftarrow \lambda {\eta }$
  \ENDFOR
 \end{algorithmic} 
 \end{algorithm}

\subsection{Training algorithm}
A sample of the training algorithm of PA-Net is presented as Algorithm~\ref{algorithm1}, where details like batch normalization and pooling layers are omitted. SGD with momentum or ADAM \cite{kingma2014adam} optimizer can be used to update parameters. Since our PA scheme approximates full precision weights and activations, using pre-trained models serves as initialization.

\subsection{Inference architecture}
Regarding the inference implementation of PA-Net, the latency is one of the most important metrics to be considered. Fortunately, the piecewise approximated weights or activations can be viewed as a linear combination of multiple binary bases ($+1$ and $0$), which indicates a parallel inference architecture.

In the forward propagation, the approximated weights are represented as follows. 
\begin{equation}
\overline W  = \sum\limits_{i = 1}^{M} {{\alpha _i}{T_i}}
\end{equation}
where ${T_i}$ is a binary weight base, given as
\begin{equation}
{T_i} = \left\{ \begin{array}{l}
\left\{ \begin{array}{l}
{B_W},W_j \in [ - \infty ,{u_1}]\\
0.0 \times {B_W},W_j \notin [ - \infty ,{u_1}]
\end{array} \right.,i = 1\\
\left\{ \begin{array}{l}
{B_W},W_j \in [{u_{i - 1}},{u_i}]\\
0.0 \times {B_W},W_j \notin [{u_{i - 1}},{u_i}]
\end{array} \right., i \in [2,\frac{M}{2}]\\
\left\{ \begin{array}{l}
{B_W},W_j \in [{u_i},{u_{i + 1}}]\\
0.0 \times {B_W},W_j \notin [{u_i},{u_{i + 1}}]
\end{array} \right., \\
~~~~~~~~~~~~~~~~~~~~~~~~~~~~~~~~~~~~~~~~~~i \in [\frac{M}{2} + 1,M - 1]\\
\left\{ \begin{array}{l}
{B_W},W_j \in [{u_M}, + \infty ]\\
0.0 \times {B_W},W_j \notin [{u_M}, + \infty ]
\end{array} \right.,i = M
\end{array} \right.
\end{equation}

Similarly, the approximated activations in the forward propagation are expressed as follows.
\begin{equation}
\overline A  = \sum\limits_{i = 1}^N {{\beta _i}{V_i}} 
\end{equation}
where ${V_i}$ is a binary activation base, given as
\begin{equation}
{V_i} = \left\{ \begin{array}{l}
\left\{ \begin{array}{l}
{B_A},A_j \in [{v_i},{v_{i + 1}}]\\
0.0 \times {B_A},A_j \notin [{v_i},{v_{i + 1}}]
\end{array} \right., i = 1,...,N - 1\\
\left\{ \begin{array}{l}
{B_A},A_j \in \left[ {{v_N}, + \infty } \right]\\
0.0 \times {B_A},A_j \notin \left[ {{v_N}, + \infty } \right]
\end{array} \right., i = N
\end{array} \right.
\end{equation}

Combined with the approximated weights, the forward propagation of the real-valued convolution can be approximated by computing $M \times N$ parallel bitwise convolutions. It is worth to notice that ${\alpha _i}{\beta _j}$ will be merged as one new scaling coefficient ${{\phi_k}}$ during the inference stage so that we omit their multiplication.   
\begin{equation}
\begin{aligned}
&\mathit{Conv}(W,A)
\approx \mathit{Conv}(\overline W,\overline A) = \mathit{Conv}(\sum\limits_{i = 1}^{M} {{\alpha _i}{T_i}} ,\sum\limits_{j = 1}^N {{\beta _j}{V_j}} ) \\
&= \sum\limits_{i = 1}^{M} {\sum\limits_{j = 1}^N {{\alpha _i}{\beta _j}} } \mathit{Conv}({T_i},{V_j}) =  \sum\limits_{k = 1}^{M \times N} {{\phi _k}Conv({T_i},{V_j})} 
\label{*}
\end{aligned}
\end{equation}

\begin{figure}
\begin{center}
   \includegraphics[width=0.95\linewidth]{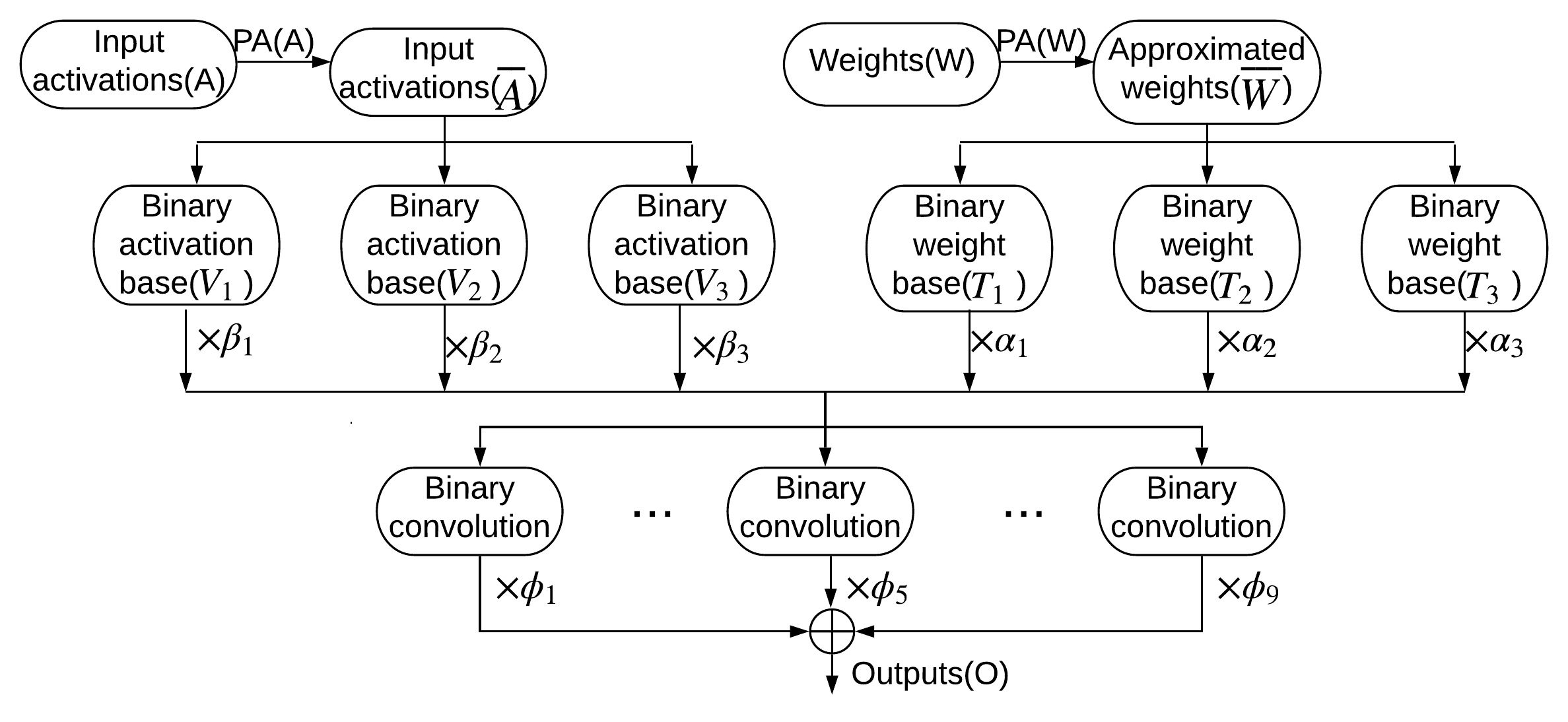}
\end{center}
   \caption{Parallel inference architecture of convolution in PA-Net}
\label{figure:inference architecture}
\end{figure}

Taking $M = 3$ and $N = 3$ as an example, both the weights and activations use $3$ bits to approximated their full precision counterpart. A full precision convolution can be computed with $9$ parallel bitwise operations and $3$ comparators, as shown in Figure~\ref{figure:inference architecture}, where the latency cost is as small as that of single binary CNNs. On the left is the structure of the activations approximation using binary activation bases ${V_1}$, ${V_2}$, and ${V_3}$. On the right is the structure of the weights approximation using binary weight bases ${T_1}$, ${T_2}$, and ${T_3}$. Thus, we implement the overall block structure of the convolution in the PA scheme with $9$ parallel bitwise operations. It is worth to notice that computing the binary convolution blocks in this figure can be directly completed by AND and popcount operations, and the binary convolution blocks do not consist of Batch Normalization or Relu layer. 

\subsection{Efficiency analysis of different binary values}
To the best of our knowledge, this is the first time to use binary values $+1$ and $0$ instead of binary values $-1$ and $+1$ for single or multiple binary CNNs, and we present their efficiency analysis in terms of costumed hardware FPGA/ASIC.

When binary convolutions are computed by bitwise operation with binary values $0$ and $+1$, the dot product of two bit-vectors $x$ and $y$ is computed using bitwise operations as follows.

\begin{equation}
\begin{aligned}
x \cdot y = \text{bitcount}(\text{AND}(x,y)),x{}_i,{y_i} \in \{ 0,+1\} {\forall _i}
\end{aligned}
\end{equation}

where $bitcount$ counts the number of bits in a bit-vector.

Similarly, when binary convolutions are computed by bitwise operation with binary values $-1$ and $+1$, the dot product of two bit-vectors $x$ and $y$ is computed using bitwise operations as follows.

\begin{equation}
\begin{aligned}
x \cdot y = N - 2 \times \text{bitcount}(\text{XNOR}(x,y)),x{}_i,{y_i} \in \{  - 1,+1\} {\forall _i}
\end{aligned}
\end{equation}

where $N$ is the number of bits in a bit-vector.

\begin{table}
\caption{2-input $7$-nm CMOS gates propagation delay, area, and power}
\label{table:gates}
\centering
\begin{tabular}{l|l|l|l}
\hline
Items & Propagation delay [$ps$] & Active area [$nm^2$] & Power [$nW$]\\
\hline
XNOR & $10.87$  & $2.90 \times 10^3$ & $1.23 \times 10^3$ \\
AND & $9.62$  & $1.45 \times 10^3$  & $6.24 \times 10^2$  \\
\hline
\end{tabular}
\end{table}

In Table~\ref{table:gates}, we present the area footprint, the input to output propagation delay and the power consumption for 2-input Boolean gates using a commercial $7$-nm FinFET technology (supply voltage $V_{DD} = 0.7V$). The active area and power consumption cost of an XNOR gate are two times as large as those of an AND gate, which indicates that the area and power consumption cost of a binary convolution with binary values $-1$ and $+1$ are two times as large as those of a binary convolution with binary values $0$ and $+1$ (except for bitcount operation).

\section{Experimental results on ImageNet dataset}
We first trained and evaluated ResNet \cite{he2016deep} using our proposed PA scheme on ImageNet ILSVRC2012 classification dataset \cite{russakovsky2015imagenet}. Then we generalize our scheme to other CNN architectures such as DenseNet and MobileNet. Finally, the computational complexity of PA-Net is analyzed on CPUs and customized hardware. 

We set the batch size of all our implementations to $64$ due to the limit on available time and resources, which slightly limits the accuracy of the results. However, the accuracy is expected to increase with a larger batch size. 

\subsection{Weights and activations approximations}
\begin{figure}[h]
\begin{subfigure}{0.23\textwidth}
\centering
\includegraphics[width=1.0\linewidth]{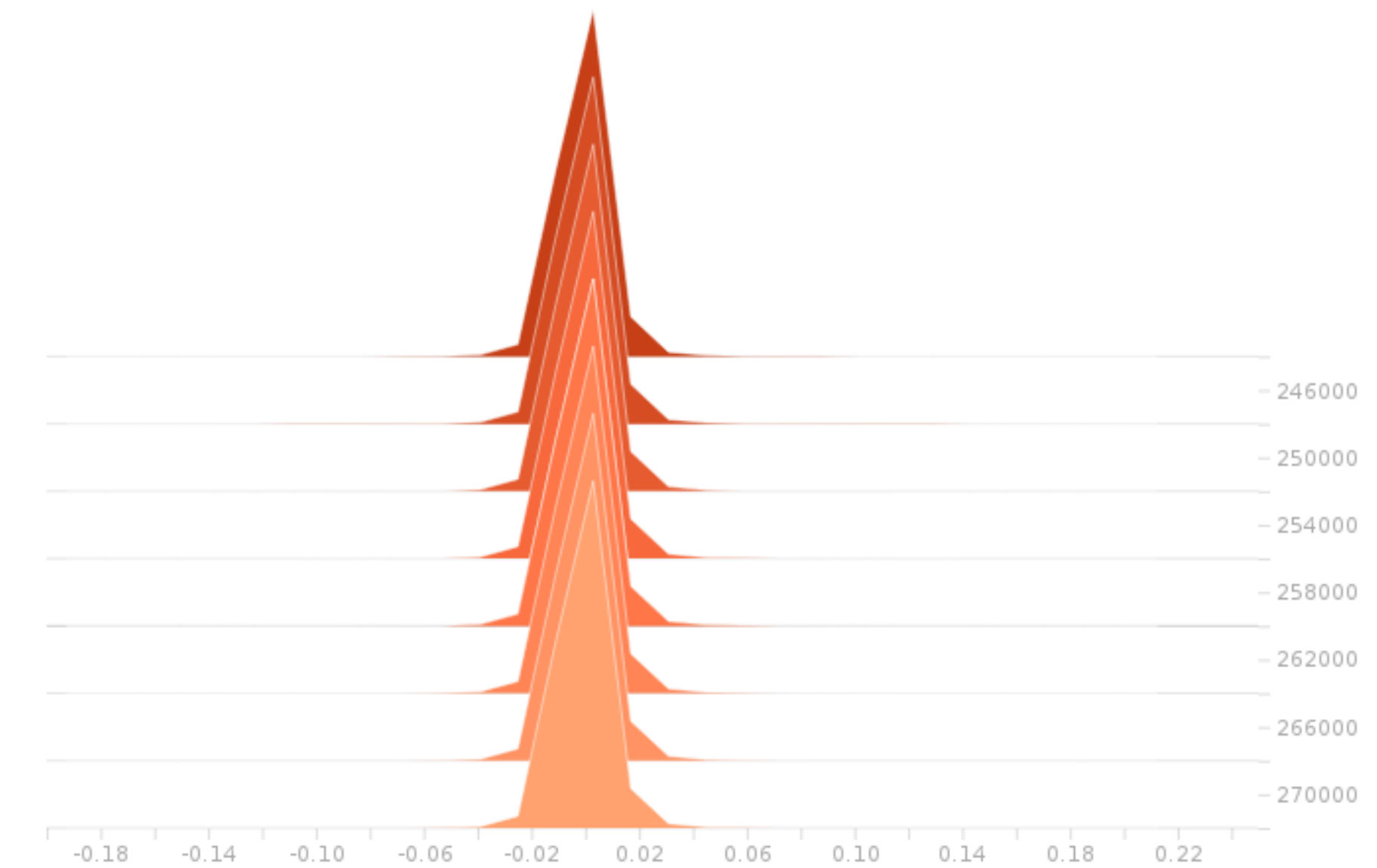}
\caption{Full precision weights} 
\label{fig3:a}
\end{subfigure}
\hspace*{\fill} 
\begin{subfigure}{0.23\textwidth}
\centering
\includegraphics[width=1.0\linewidth]{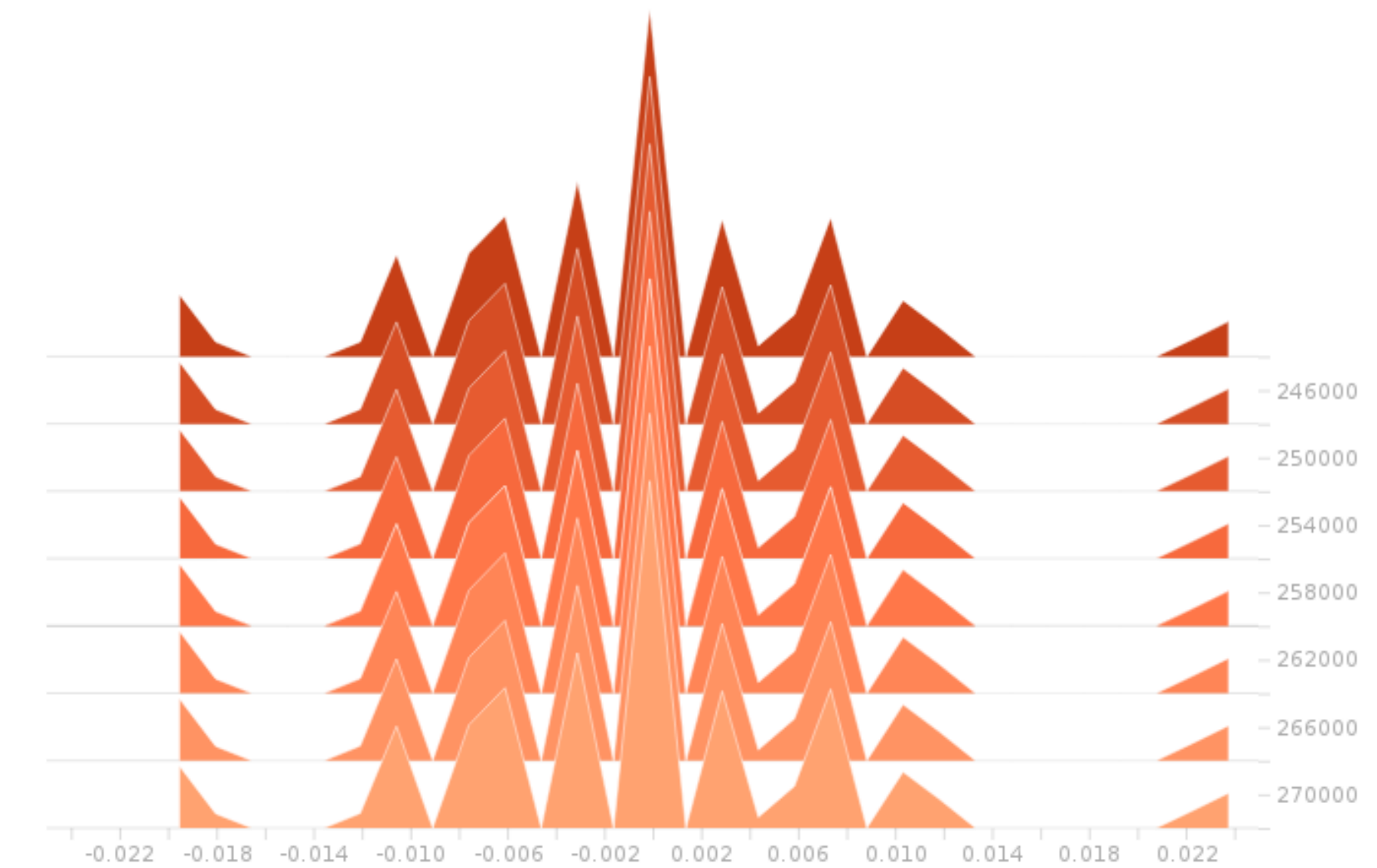}
\caption{Approximated weights} 
\label{fig3:b}
\end{subfigure}
\caption{Distribution of full precision and approximated weights.} 
\label{fig3}
\end{figure}

\begin{figure}[h]
\begin{subfigure}{0.23\textwidth}
\centering
\includegraphics[width=1.0\linewidth]{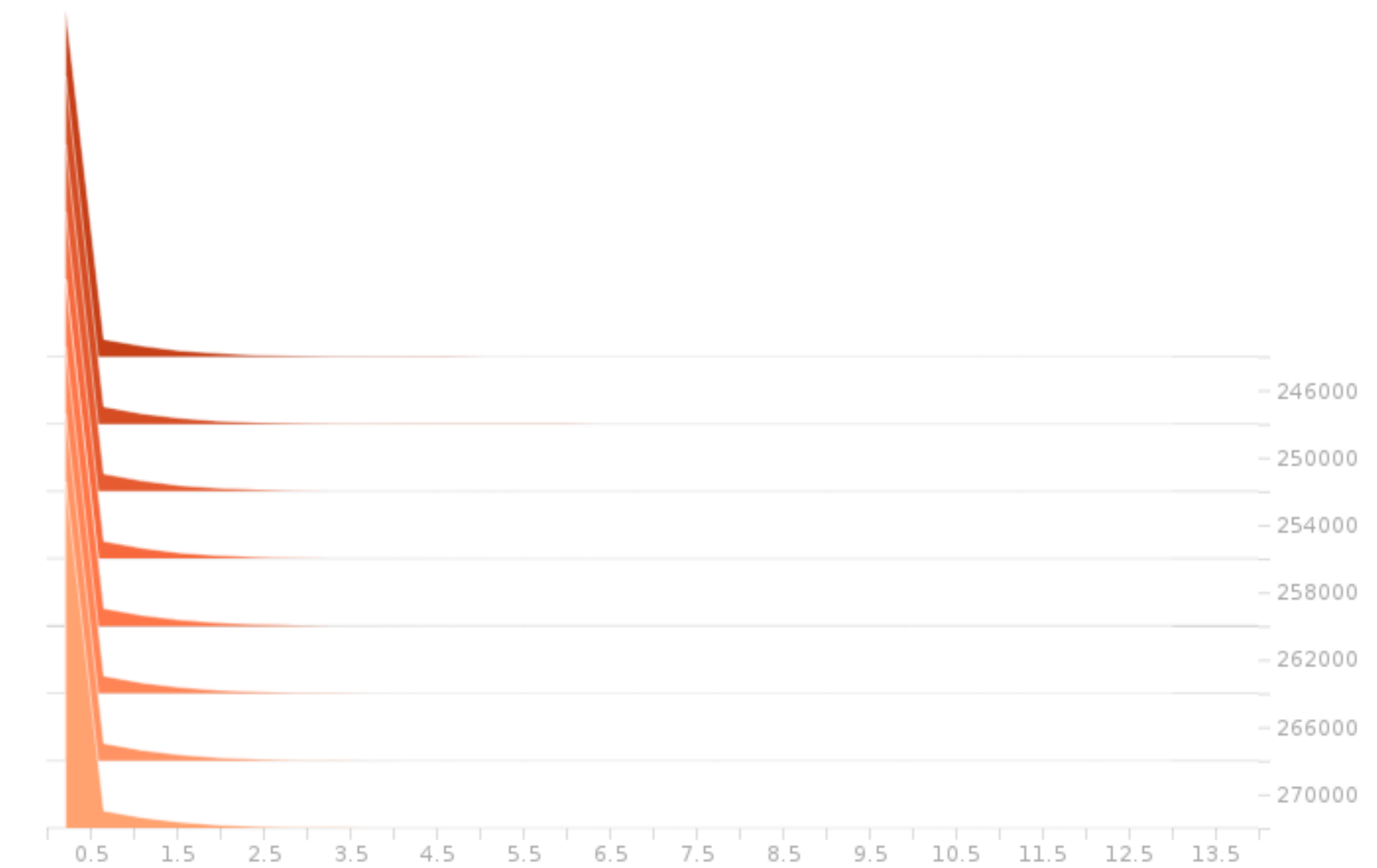}
\caption{Full precision activations} 
\label{fig4:a}
\end{subfigure}
\hspace*{\fill} 
\begin{subfigure}{0.23\textwidth}
\centering
\includegraphics[width=1.0\linewidth]{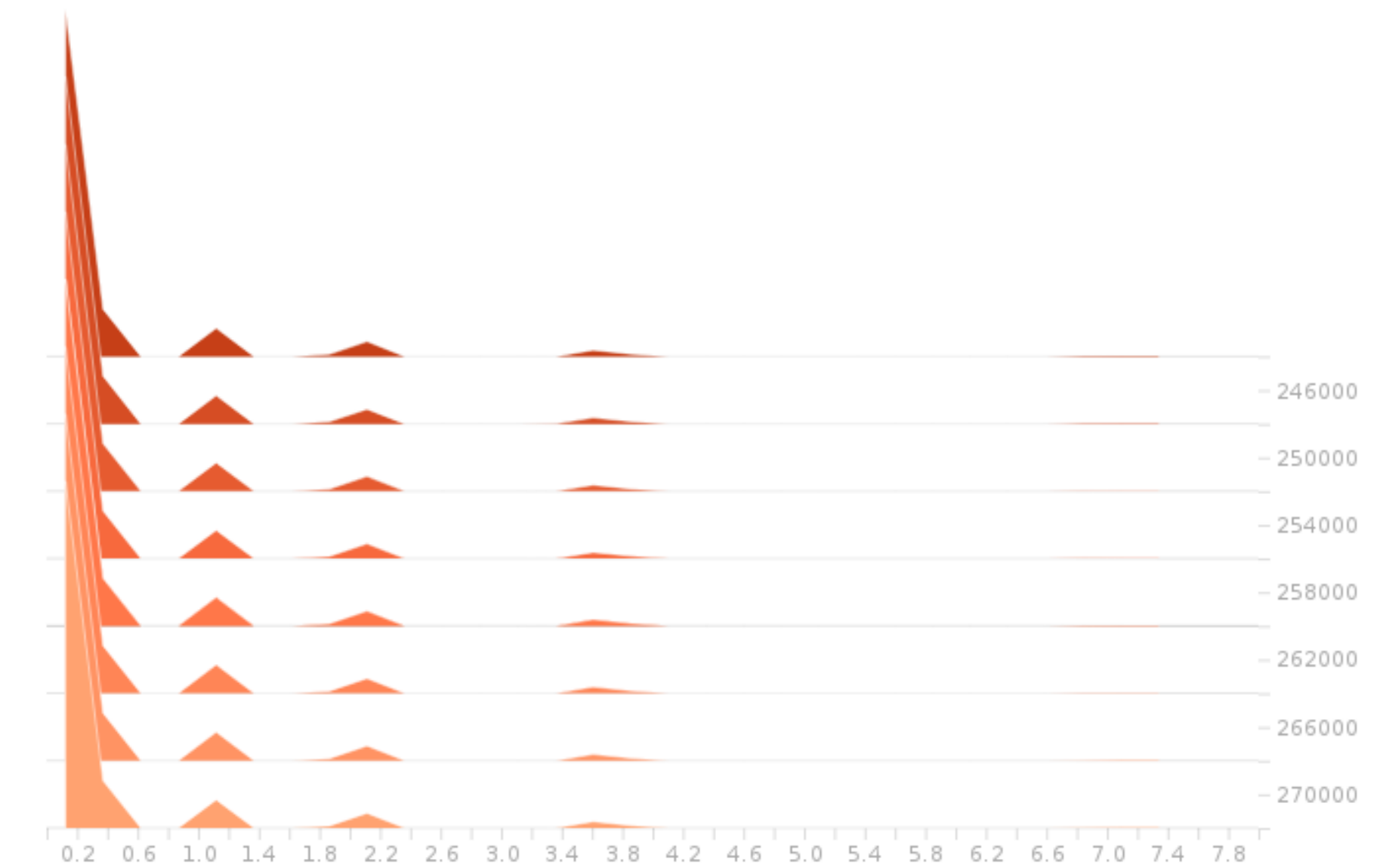}
\caption{Approximated activations} 
\label{fig4:b}
\end{subfigure}
\caption{Distribution of full precision and approximated activations.} 
\label{fig4}
\end{figure}

Using the ResNet18/group2/block1/conv1 layer, we sampled full precision weights and their approximations with $M = 8$. Their histograms are shown in Figure~\ref{fig3:a} and \ref{fig3:b}, respectively. Horizontal axis and longitudinal axis represent the values and the number of values of weights/activations, respectively. Similarly, the comparison of activation histograms are shown in Figure~\ref{fig4}, which are acquired from the ResNet18/group2/block1/conv2 layer and include the full precision activations in Figure~\ref{fig4:a} and their approximation with $N  = 5$ in Figure~\ref{fig4:b}. As the comparisons show, the distributions of the approximated weights and activations are similar to those of the full precision weights and activations, respectively, which means that PA scheme provides an accurate way for multiple binary bases to approximate the distribution of their full precision counterparts.

\subsection{Comparison with ABC-Net}
Both PA-Net and ABC-Net can utilize parallel bitwise operation and achieve higher accuracy than single binary CNNs, so the differences between them need to be analyzed. The accuracy comparisons between PA-Net and ABC-Net are shown in Table~\ref{table:comparison with abc}. 

\begin{table*}
\caption{Comparison with ABC-Net using ResNet as backbones}
\label{table:comparison with abc}
\centering
\begin{tabular}{lllllll}
\hline
 Model & $M$ & $N$ & ${Top-1}$ & ${Top-5}$ & $Top-1~gap$ & $Top-5~gap$ \\
\hline
ABC-ResNet18  & $5$  & full precision & $68.3\%$ & $87.9\%$ & $1.0\%$ & $1.3\%$\\
PA-ResNet18  & $4$  & full precision & $68.4\%$ & $88.3\%$ & $0.9\%$ & $0.9\%$\\
PA-ResNet18  & $8$  & full precision & $69.3\%$ & $88.9\%$ & $0.0\%$ & $0.3\%$\\
\hline
ABC-ResNet18 & $5$  & $5$ & $65.0\%$ & $85.9\%$ & $4.3\%$ & $3.3\%$\\
PA-ResNet18 & $4$  & $5$ & $66.6\%$ & $87.1\%$ & $2.7\%$ & $2.1\%$\\
PA-ResNet18 & $8$  & $7$ & $68.1\%$ & $88.1\%$ & $1.2\%$ & $1.1\%$\\
ResNet18  &full precision  &full precision  & $69.3\%$ & $89.2\%$ & $-$ & $-$\\
\hline
ABC-ResNet34 & $5$  & $5$ & $68.4\%$ & $88.2\%$ & $4.9\%$ & $3.1\%$\\
PA-ResNet34 & $4$  & $5$ & $70.1\%$ & $89.2\%$ & $3.2\%$ & $2.1\%$\\
PA-ResNet34 & $8$  & $7$ & $71.5\%$ & $90.0\%$ & $1.8\%$ & $1.3\%$\\
ResNet34  &full precision  &full precision  & $73.3\%$ & $91.3\%$ & $-$ & $-$\\
\hline
ABC-ResNet50 & $5$  & $5$ & $70.1\%$ & $89.7\%$ & $6.0\%$ & $3.1\%$\\
PA-ResNet50 & $4$  & $5$ & $73.0\%$ & $91.0\%$ & $3.1\%$ & $1.8\%$\\
PA-ResNet50 & $8$  & $7$ & $74.3\%$ & $91.9\%$ & $1.8\%$ & $0.9\%$\\
ResNet50  &full precision  &full precision  & $76.1\%$ & $92.8\%$ & $-$ & $-$\\
\hline
\end{tabular}
\end{table*}

\begin{table*}
\caption{Generalization to DenseNet and MobileNet.}
\label{table:generalization of PA scheme}
\centering
\begin{tabular}{lllllll}
\hline
 Model & $M$ & $N$ & $Top-1$ & $Top-5$ & $Top-1~gap$ & $Top-5~gap$ \\
\hline
PA-DenseNet121 & $8$  & $6$ & $72.3\%$ & $90.8\%$ & $2.7\%$ & $1.5\%$\\
DenseNet121  &full precision  &full precision  & $75.0\%$ & $92.3\%$ & $-$ & $-$\\
\hline
PA-1.0 MobileNet-224 & $8$  & $7$ & $69.0\%$ & $88.4\%$ & $1.6\%$ & $1.5\%$\\
1.0 MobileNet-224  &full precision  &full precision  & $70.6\%$ & $89.9\%$ & $-$ & $-$\\
\hline
\end{tabular}
\end{table*}

Table~\ref{table:comparison with abc} shows that PA-Net achieves higher accuracy than ABC-Net while requiring less overhead, which strongly supports the idea that PA-Net provides a better approximation than ABC-Net for both the weights and activations. In addition, Table~\ref{table:comparison with abc} shows the unique advantage of PA-Net over ABC-Net since PA-Net can give higher accuracy for multiple binary CNNs by increasing $M$ and $N$. However, we also re-implemented ABC-Net and reproduced the results, which shows that its accuracy remains unchanged (or even becomes worse) as we keep increasing $M$ and $N$ more than $5$. 

For the weights approximation only (i.e., when $N$ is full precision), PA-ResNet18 gives no Top-1 accuracy loss with $M = 8$. PA-ResNet achieves higher accuracy with $M = 4$ and $N = 5$ than ABC-ResNet with $M = 5$ and $N = 5$, which means that PA-ResNet provides better approximation with less overhead. PA-ResNet with $M = 8$ and $N = 7$ reduce the Top-5 accuracy gap around $1.0\%$. But the accuracy of ABC-Net remains unchanged or even becomes worse with the increase of $M$ and $N$ more than $5$ based on our re-implementation. PA-Net is expected to reach no accuracy loss with the increase of $M$ and $N$, which we have not attempted due to the limitations of computational resources, training time, and the slow increase trend of accuracy with the increase of $M$ and $N$.

\subsection{Generalization to other CNN architectures}
To demonstrate the generalization of PA scheme, we applied it on 1.0 MobileNet-224 \cite{howard2017mobilenets} and DenseNet121 \cite{huang2017densely}. The results are shown in  Table~\ref{table:generalization of PA scheme}. Due to memory limitation, we implemented PA-DenseNet121 with $N = 6$. Its Top-1 accuracy loss is $2.7\%$, which is expected to decrease further with increasing $N$. Top-1 accuracy loss of PA-1.0 MobileNet-224 achieves $1.6\%$ with $N = 7$. Pointwise convolution is binarized while depthwise convolution is kept as full precision convolution since they do not need significant computational resources.

\subsection{Generalization to object detection}
We choose SSD300 with the backbone network of ResNet50 as our baseline. The training dataset is VOC2007 + 2012, while the testing dataset is VOC2007~\cite{Everingham15}. In the SSD300 model, we use the layers from Conv1 to Conv5\_x of the pre-trained ResNet50 as the backbone network, apply residual blocks as the extra layers, and keep the number of feature maps the same as the original implementation~\cite{SSD}. All the backbone layers except Conv1 are binarized, while all the convolutional layers of the head network remain in full precision. We train the full precision ResNet50 on the ImageNet classification dataset as the backbone network, and then the full precision object detector SSD300 using the pre-trained ResNet50. Finally, we binarize and finetune the pre-trained object detector SSD300 with the PA scheme.

\begin{table}
\caption{Performances of the full-precision SSD300 network and its binary counterpart.}
    \label{tab:ssd}
    \centering

    \begin{tabular}{l|l|l|l|l}
         \hline
          \text{Detector}& \text{Backbone} &\text{Weights}
          &\text{Activations}
          & \text{mAP@0.5} \\
         \hline
         SSD300 & ResNet50 & Full precision & Full precision  & 74.35 \\
         SSD300 & ResNet50 &$M = 4$ &Full precision  & 72.53 \\
         SSD300 & ResNet50 &$M = 4$ & $N = 5$  & 58.60 \\
         \hline
    \end{tabular}
\end{table}

Applying the PA scheme to the SSD300 network, we present the results in Table~\ref{tab:ssd}.  When only weights are binarized using the PA scheme with $M=4$, the binary SSD300 model achieves comparable accuracy by $1.82$ mAP reduction compared with its full precision baseline networks. When applying the PA scheme with binary weights ($M = 4$) and binary activations ($N = 5$)  for the SSD300 network, the binary SSD300 network shows an accuracy reduction in $15.75$ mAP, which outperforms the real-time full-precision Fast YOLO~\cite{DBLP:yolo} ($52.7$ mAP).

\subsection{Comparisons with state-of-the-art methods}

\begin{table}
\caption{Accuracy comparisons of ResNet18 with different quantized methods.}
\label{table:comparison with STA}
\begin{center}
\begin{tabular}{lllll}
\hline
Model & $W$ & $A$ & $Top-1$ & $Top-5$  \\
\hline
Full Precision  & $32$  & $32$ & $69.3\%$ & $89.2\%$  \\
\hline
BWN  & $1$  & $32$ & $60.8\%$ & $83.0\%$ \\
XNOR-Net   & $1$  & $1$ & $51.2\%$ & $73.2\%$ \\
Bi-Real Net  & $1$  & $1$ & $56.4\%$ & $79.5\%$ \\\hline
ABC-Net ($M = 5$, $N = 5$) & $1$  & $1$ & $65.0\%$ & $85.9\%$ \\
Group-Net ($5$ bases) & $1$  & $1$ & $64.8\%$ & $85.7\%$ \\
\hline
DoReFa-Net & $2$ & $2$ & $62.6\%$ & $84.4\%$ \\
SYQ  & $1$ & $8$ & $62.9\%$ & $84.6\%$ \\
LQ-Net & $2$ & $2$ & $64.9\%$ & $85.9\%$ \\ 
\hline
PA-Net ($M = 8$, $N = 7$)  & $1$  & $1$  & $68.1\%$  & $88.0\%$ \\
PA-Net ($M = 8$) & $1$  & $32$ & $69.3\%$ & $88.9\%$ \\
\hline
\end{tabular}
\end{center}
\end{table}

\begin{table*}
\caption{Memory usage and Flops calculation of Bi-Real Net, PA-Net, and full precision models.}
\label{table:analysis}
\centering
\begin{tabular}{lllll}
\hline
 Model & Memory usage & Memory saving & Flops & Speedup  \\
\hline
Bi-Real ResNet18  &$ 33.6$Mbit  &$11.14~\times$  & $ 1.67 \times 10^8
$ & $10.86~\times$ \\
ABC-ResNet18 & $77.1$Mbit  & $4.85~\times$ & $6.74 \times 10^8$ & $2.70~\times$ \\
PA-ResNet18 & $61.6$Mbit  & $6.08~\times$ & $6.74 \times 10^8$ & $2.70~\times$ \\
ResNet18  &$ 374.1$Mbit  &$-$  & $ 1.81 \times 10^9
$ & $-$ \\
\hline
Bi-Real ResNet34  &$ 43.7$Mbit  &$15.97~\times$  & $ 1.81 \times 10^8
$ & $18.99~\times$ \\
ABC-ResNet34 & $106.3$Mbit  & $6.56~\times$ & $1.27 \times 10^9$ & $2.88~\times$\\
PA-ResNet34 & $85.0$Mbit  & $8.20~\times$ & $1.27 \times 10^9$ & $2.88~\times$\\
ResNet34  &$ 697.3$Mbit  &$-$  & $3.66 \times 10^9$ & $-$\\
\hline
Bi-Real ResNet50 & $176.8$Mbit  & $4.62~\times$ & $5.45 \times 10^8$ & $7.08~\times$\\
ABC-ResNet50 & $201.6$Mbit  & $4.06~\times$ & $1.44 \times 10^9$ & $2.68~\times$\\
PA-ResNet50 & $161.3$Mbit  & $5.07~\times$ & $1.44 \times 10^9$ & $2.68~\times$\\
ResNet50  &$817.8$Mbit  &$-$  & $3.86 \times 10^9$ & $-$\\
\hline
\end{tabular}
\end{table*}

\begin{table*}
\caption{Latency cost of Bi-Real Net, PA-Net, and full precision models. ${T_{XNOR}}$, ${T_{pop}}$, ${T_{mul}}$, ${T_{AND}}$, ${T_{com}}$, ${T_{add}}$ refer to the delay time of a XNOR, popcount, multiplication, AND, comparison, and addition operation, respectively.}
\label{table:latency}
\centering
\begin{tabular}{lll}
\hline
 Model & Latency cost  & Speedup \\
\hline
Bi-Real Net  &${c_{in}}hw \times ({T_{XNOR}} + {T_{pop}}) + {T_{mul}}$  &$\approx ({T_{mul}} + {T_{add}})/({T_{XNOR}} + {T_{pop}})$  \\
PA-Net & ${c_{in}}hw \times ({T_{AND}} + {T_{pop}}) + 5{T_{mul}} + 4{T_{add}} + {T_{com}}$ & $\approx ({T_{mul}} + {T_{add}})/({T_{AND}} + {T_{pop}})$    \\
Full precision models  &${c_{in}}hw \times {T_{mul}} + ({c_{in}}hw - 1) \times {T_{add}}$ & $-$ \\
\hline
\end{tabular}
\end{table*}

The comparisons between PA-Net and recent developments are shown in Table~\ref{table:comparison with STA}, where PA-Net adopts the configuration of $M=8$ and $N=7$. Regarding single binary models BWN, XNOR-Net \cite{rastegari2016xnor} and Bi-Real Net \cite{liu2018bi}, and multiple parallel binary models ABC-Net\cite{lin2017towards} and Group-Net\cite{zhuang2019structured}, PA-Net outperforms them by much higher accuracy. When it comes to the comparison with fix-point quantization DoreFa-Net \cite{zhou2016dorefa,Zhang_2018_ECCV,faraone2018syq}, fixed-point CNNs can achieve the same or even higher performance with carefully customized bit-widths than PA-Net. But the advantage of PA-Net is the parallelism of inference architecture, which provides a much lower latency using bitwise operation than fixed-point CNNs.

\subsection{Computational complexity analysis}
In this part, we analyze and compare the computational complexity of Bi-Real Net (Liu et al. 2018), PA-Net, and full precision models
on current CPUs in terms of computation and memory usage, and on customized hardware (i.e., FPGA/ASIC) in terms of latency. Bi-Real
Net maintains high efficiency and achieves the state-of-the-art
accuracy as a single binary CNN. During this analysis, PA scheme uses 4 bases for weights and 5 bases for activations
approximation. 

\subsubsection{Computation and memory usage analysis}
We analyze and compare the computational complexity of Bi-Real Net \cite{liu2018bi}, PA-Net and full precision models, and their memory saving and speedup are shown in Table~\ref{table:analysis}.

Unlike full precision models which require real-valued parameters and operations, PA-Net and Bi-Real Net have binary and real-valued parameters mixed, so their execution requires both bitwise and real-valued operations. To compute the memory usage of PA-Net and Bi-real Net, we use $32$ bit times the number of real-valued parameters and $1$ bit times the number of binary values, which are summed together to get their total number bit. We use Flops as the main metrics to measure the bitwise operations, the real-valued operations, and the speedup of implementation. Since the current generation of CPUs can compute bitwise AND and popcount operations in parallelism of $64$, the Flops to compute PA-Net and Bi-Real Net is equal to the number of the real-valued multiplications, comparisons, and $1/64$ of the number of the bitwise operations.

We follow the suggestion from \cite{rastegari2016xnor,liu2018bi} to keep the weights and activations of the first convolutional and the last fully connected layer as real-valued. It is worthy to notice that we binarize all the $1 \times 1$ downsampling layer in PA-Net to further reduce the computational complexity.

For ResNet18, ResNet34, and ResNet50, our PA scheme can reduce memory usage by more than 5 times and achieves a computation reduction of nearly 3 times, in comparison with the full precision counterpart. Compared with Bi-Real ResNet50, the computation reduction of our proposed PA-ResNet50 with $4$ weights bases and $5$ activations bases is only two times smaller, and it even requires less memory usage because of the binarization of the downsampling layer. 

Combining Table 3 and Table 7, we can conclude that PA-Net can achieve better accuracy ($1.6\%$, $1.7\%$, and $2.9\%$ for ResNet18, ResNet34, and ResNet50) while consuming fewer parameters ($15.4$Mbit, $21.3$Mbit, and $40.33$Mbit for ResNet18, ResNet34, and ResNet50) and the same Flops compared to ABC-Net during the inference stage.

\subsubsection{Latency analysis} 
To be implemented on customized hardware (i.e., FPGA/ASIC), latency cost is one of the most important metrics for real-time applications. As shown in Table~\ref{table:latency}, the latency cost of an individual convolution in Bi-Real Net, PA-Net, and full precision models is analyzed, where we assume that the convolution implementation is paralleled thoroughly. Compared with full precision models, the latency cost of PA-Net and Bi-Real Net is significantly reduced. ${T_{AND}}$ is smaller than ${T_{XNOR}}$, and the latency cost of a convolution in PA-Net increased only by $4T_{mul} + 4T_{add} + T_{com}$ compared with that in Bi-Real Net.

\section{Conclusions}

In this paper, we introduced the PA scheme for multiple binary CNNs, which adopts piecewise functions for both the forward propagation and backpropagation. Compared with state-of-the-art single and multiple binary CNNs, our scheme provides a better approximation for both full precision weights and activations. We implemented our scheme over several modern CNN architectures, such as ResNet, DenseNet, and MobileNet, and tested on classification task using ImageNet dataset. Results are competitive and almost close the accuracy gap compared with their full precision counterparts. Because of the binarization of downsampling layer, our proposed PA-ResNet50 requires less memory usage and only two times Flops than Bi-Real Net with $4$ weights and $5$ activations bases, which shows its potential efficiency advantage over single binary CNNs with a deeper network.

\clearpage

\bibliography{ecai}
\end{document}